\documentclass[11pt,a4paper]{article}
\usepackage[hyperref]{emnlp2020}
\usepackage{times}
\usepackage{latexsym}
\usepackage{amsmath}
\usepackage{graphicx}
\usepackage{booktabs}
\usepackage{amsfonts}
\usepackage{soul}

\usepackage{multirow}

\newcommand{\todo}[1]{}
\renewcommand{\todo}[1]{{\color{red} TODO. {#1}}}

\usepackage{microtype}

\aclfinalcopy 

\title{Exploring Neural Entity Representations for Semantic Information}

\author{Andrew Runge \\
    Duolingo \\
    Pittsburgh, PA, USA \\
  \texttt{arunge@duolingo.com} \\\And
  Eduard Hovy \\
  Carnegie Mellon University \\
  Pittsburgh, PA, USA \\
  \texttt{hovy@cmu.edu} \\}

\date{August 15, 2020}

\begin{document}
\maketitle
\begin{abstract}

Neural methods for embedding entities are typically extrinsically evaluated on downstream tasks and, more recently, intrinsically using probing tasks.
Downstream task-based comparisons are often difficult to interpret due to differences in task structure, while probing task evaluations often look at only a few attributes and models.
We address both of these issues by evaluating a diverse set of eight neural entity embedding methods on a set of simple probing tasks, demonstrating which methods are able to remember words used to describe entities, learn type, relationship and factual information, and identify how frequently an entity is mentioned.
We also compare these methods in a unified framework on two entity linking tasks and discuss how they generalize to different model architectures and datasets.

\end{abstract}

\section{Introduction}

Neural methods for generating entity embeddings have become the dominant approach to representing entities, with embeddings learned through methods such as pretraining, task-based training, and encoding knowledge graphs \cite{yamada2016words,ling2020relic,wang2020kepler}. 
These embeddings can be compared extrinsically by performance on a downstream task, such as entity linking (EL).
However, performance depends on several factors, such as the architecture of the model they are used in and how the data is preprocessed, making direct comparison of the embeddings hard.

Another way to compare these embeddings is intrinsically using probing tasks \cite{yaghoobzadeh2016intrinsic,conneau2018vector}, which have been used to examine entity embeddings for information such as an entity's type, relation to other entities, and factual information \cite{yaghoobzadeh2017multi,peters2019know,petroni2019language,ling2020relic}.
These prior examinations have often examined only a few methods, and some propose tasks that can only be applied to certain classes of embeddings, such as those produced from a mention of an entity in context.

We address these gaps by comparing a wide range of entity embedding methods for semantic information using both probing tasks as well as downstream task performance.
We propose a set of probing tasks derived simply from Wikipedia and DBPedia, which can be applied to any method that produces a single embedding per entity.
We use these to compare eight entity embedding methods based on a diverse set of model architectures, learning objectives, and knowledge sources. 
We evaluate how these differences are reflected in performance on predicting information like entity types, relationships, and context words.
We find that type information is extremely well encoded by most methods and that this can lead to inflated performance on other probing tasks.
We propose a method to counteract this and show that it allows a more reliable estimate of the encoded information.
Finally, we evaluate the embeddings on two EL tasks to directly compare their performance when used in different model architectures, identifying some that generalize well across multiple architectures and others that perform particularly well on one task.

We aim to provide a clear comparison of the strengths and weaknesses of various entity embedding methods and the information they encode to guide future work.
Our probing task datasets, embeddings, and code are available online.\footnote{https://github.com/AJRunge523/entitylens}

\section{Models} \label{sec:models}

We compare eight different approaches to generating entity embeddings, organized along two dimensions: the training process of the underlying model, and the content used to inform the embeddings.

Along the training dimension, the first method is \textbf{task-learned embeddings}, which are learned as part of a downstream task, such as EL. 
\textbf{Pretrained embeddings} are learned through a dedicated pretraining phase designed to produce entity embeddings. Finally, \textbf{derived embeddings} are produced by models capable of embedding any generic text, but that had no specific entity-based training.

Along the content dimension, the first type is \textbf{description-based embeddings}, which are learned or generated from a text description of the entity.
\textbf{Context-based embeddings} are learned from words surrounding mentions of the entity.
Lastly, \textbf{graph-based embeddings} are learned entirely from a knowledge graph, linking entities to types and to each other.
Models may leverage multiple types of information to learn embeddings.

We use the March 5, 2016 dump of Wikipedia to train our task-learned and pretrained embedding models, while the derived embedding models are publicly available pre-trained language models.\footnote{https://code.google.com/archive/p/word2vec}\footnote{https://huggingface.co/bert-base-uncased}\footnote{https://huggingface.co/bert-large-uncased}

\subsection{Task-Learned Embedding Models}

For our task-learned models, we re-implement two neural EL models, which learn entity representations for the goal of connecting mentions of entities in text to entities in a knowledge base (KB). 
We briefly summarize them here and refer interested readers to the original papers for further details.

First is the \textbf{CNN}-based model of \citet{francislandau2016convolutional}, a description and context-based hybrid model.
It encodes text mentions of entities by applying convolutions over the mention's name, context sentence, and the first 500 words of the document it appears in and encodes candidate KB entities with convolutions over the entity's name and first 500 words of its Wikipedia page.
It computes cosine distance between the outputs of each of the mention and KB convolutions, producing six features which are passed to a linear layer to produce a score for each candidate, trying to maximize the score of the true candidate.
We use a kernel size of 150 and concatenate the candidate name and document convolution outputs to get 300-dimensional entity embeddings from this model.

Second is the \textbf{RNN}-based model of \citet{eshel2017noisy}, a context-based model which learns a 300-dimension embedding for each KB entity.
Each mention is represented by two 20-word context windows on its left and right, which are passed through single-layer bidirectional GRUs.
The RNN outputs are each passed to an MLP attention module which uses the candidate entity embedding as the attention context to pass information from the text to the embeddings.
The attention outputs and entity embedding are concatenated and passed through a single-layer MLP followed by a linear layer to compute a score for the candidate.

We train these models using an EL dataset built from all of Wikipedia  \cite{eshel2017noisy,gupta2017joint}.
We take the anchor text of each intra-Wiki link in Wikipedia as a mention, with the page it links to as the gold entity, filtering any cross-wiki links, non-entity pages, and entities with fewer than 20 words to create 93.8M training instances.
Each mention is assigned a single negative candidate randomly from all entities \cite{eshel2017noisy}.
We train each model for a single epoch on this dataset, following Eshel's method.

\subsection{Pretrained Entity Models}
We evaluate three pretrained embedding models that leverage context and graph-based information to represent entities. For all three models, we train 300 dimensional entity embeddings.

First is the context-based model of \citet{ganea2017joint} (\textbf{Ganea}).
This model learns entity representations by sampling a distribution of context words around mentions of each entity and moves the entity embeddings closer to words in the entity's context distribution and further from words sampled from a uniform distribution.
The embeddings are normalized, resulting in a joint distribution of entities and words around the unit sphere, where entity vectors are close to their context words.
We retrain their model on a larger subset of 1.5 million entities that includes the entities we use for the probing and EL tasks with a context window of 10 and 30 negative samples until it matches the authors' original scores on an entity similarity metric \cite{ceccarelli2013related}.

We next use the graph-based BigGraph model \cite{lerer2019biggraph}.
It learns entity and relationship embeddings from a knowledge graph where the relation embeddings define transformation functions between the source and target of a relationship, giving semantic meaning to the distance between entities.
We extract type and relationship triples from Wikipedia using the DBPedia toolkit \footnote{https://wiki.dbpedia.org/} and train the model on the resulting graph for 50 epochs.

Third, we use the Wikipedia2Vec toolkit of \citet{yamada2018wiki2vec}, a context and graph-based hybrid model which jointly trains word and entity embeddings (\textbf{Wiki2V}). 
It learns the word and entity embeddings using three tasks: 1) a skip-gram word prediction task, 2) an entity context task that predicts context words for each entity, and 3) an entity graph link prediction model that predicts which entities link to a given entity. 
We train the embeddings for 10 epochs, using the same context window and negative samples as the \textbf{Ganea} model.

\subsection{Derived Models}

Our first derived model is a simple bag of vectors model, in which we average the GoogleNews Word2Vec \cite{Mikolov2013word2vec} vectors of the first 512 words of the entity's Wikipedia page. 
Our other two derived models are BERT-based embeddings \cite{devlin2018bert} of the first 512 words in the entity's Wikipedia page. We use \texttt{BERT-base-uncased} and \texttt{BERT-large-uncased}, which generate 768 and 1024 dimensional embeddings for each entity by averaging all the hidden states of all tokens in the final layer.
We explored averaging the hidden states of the CLS tokens in different layers in initial experiments, but found averaging all hidden states in the final layer performed best.

\section{Entity Embedding Probing Tasks}

We next introduce a set of 22 probing tasks which can be applied to all of the embedding methods described above, divided into 5 categories based on the information they probe: \textit{context words} used to describe a given entity, \textit{entity type} information, \textit{relationships} between entities, how \textit{frequently} an entity is referenced, and \textit{factual knowledge}.

\subsection{Context Word Identification}

Except for \textbf{BigGraph}, all of our models' embeddings are trained on either text describing an entity or text surrounding mentions of an entity.
As such, we explore how well the embeddings can recognize words used in the context of a given entity.
We define an entity's context words as the words which appear at least once in both 1) the first 500 words of the entity's Wikipedia page and 2) a ten word window around an anchor link to that entity.
By ensuring each word appears both in context with an entity and in the description, we can avoid biasing the task towards context or description-based embeddings.
We create a binary prediction task for whether or not a word appears in an entity's context words for 1,000 high frequency (appearing in $>$100k Wiki pages, \textbf{W-H}) and 1,000 mid-frequency words ($>$10k, \textbf{W-M}).

\subsection{Entity Types and Sub-types} 

Similar to prior work \cite{yaghoobzadeh2017multi,chen2020type}, we examine how well different entity embedding methods are able to learn entity type information using probing tasks based on the DBPedia\footnote{https://wiki.dbpedia.org/services-resources/ontology} ontology.
We extract the types from each of the first 3 levels of the ontology, representing increasingly fine-grained entity types, and create one N-way classification task for all types at that level, which we refer to as \textbf{T-1}, \textbf{T-2}, and \textbf{T-3}.

\subsection{Relation Prediction}

We probe for relationships between entities in three ways: 1) How reliably a relation type can be identified between a pair of entities, 2) how well the type of a relationship between a pair of entities can be predicted, and 3) how well the fact that two entities are related can be detected.

\subsubsection{Binary Relation Identification}

With binary relation identification, our goal is to determine if a given relationship type can be identified reliably between pairs of entities.
We extract relationship triples for 244 relationship types between entity pairs from DBPedia and build a binary classification task for each of them (\textbf{R-I}).

DBPedia only contains positive relationship examples between head and tail entities, so to create the binary tasks we must construct negative examples. 
We create negative examples by randomly replacing either the head or the tail in a positive example, weighted by how often the head entity appears as the head for this relationship type, and similarly for the tail entity \cite{wang2014graph}.
This reduces the risk of accidentally generating false negative corrupted relationships (true relationships that weren't in DBPedia) by making it more likely that in N-to-1 or 1-to-N relationship types we replace the `1' entity.

One risk with this approach is entity type leakage.
If the replaced entity has an unlikely semantic type for the given relationship pair, entity embeddings that strongly encode type information may be able to easily detect the fake relationships based solely on the two entities' types, rather than true knowledge of their relationship.
To address this, we modify the above replacement algorithm so that when replacing entity $E$, we select a replacement $E'$ that matches the entity type to the finest grained type possible in the DBPedia ontology.
If the replaced entity has no type in the ontology, we select a random entity that has appeared in the same role (head or tail) for this relationship in the KB.

\subsubsection{Relationship Classification} 
We use the 244 extracted relationship types from the relation identification task to create a 244-way relationship classification task (\textbf{R-C}).
We also combine this dataset with the previous task, to create a 245-way relationship classification task that includes corrupted relationships for each type with the label \textit{None} (\textbf{R-C+I}). 
If the representations can detect entity types effectively, then certain types of relationships may be easier to classify based solely on the types of the entities involved.
Our type-restricted relationship corruption method should help ensure that good performance on this task requires understanding the relationships themselves rather than just the types of the involved entities.

\subsubsection{Relationship Detection}

Finally, we examine the general task of predicting whether a pair of entities is related or not, which requires an explicit relationship between two entities compared to an entity relatedness task \cite{hoffart2011kore,newman2018wikisrs}.
Effectively encoding this information can help with tasks like knowledge graph completion, where knowing the existence of a link is useful, even if the exact type of the link is unknown.
We sample a small number of positive examples and their corruptions from each of the 244 relationship types as described above to create this task (\textbf{R-D}).

\subsection{Entity Popularity}

Prior work has found that incorporating the probability of an entity being linked to in a knowledge base is useful for downstream tasks such as EL \cite{yamada2016words,eshel2017noisy}.
We define popularity as how frequently a given entity is linked to in Wikipedia from Wiki pages. 
We compute the popularity of each entity in Wikipedia and construct three types of tasks to probe for this information.
First is a regression task, predicting the log-scaled number of times an entity is linked to (\textbf{P-R}).
The second is a multi-class classification task for the binned number of times an entity is linked to as a coarser popularity estimate, with bins for $>1000$, $100-1000$, $10-100$, and $1-10$ links (\textbf{P-B}).
The third is a comparative task, where the model must predict which of two entities is linked to more often.
For fine-grained analysis, we select pairs for comparison based on the relative difference in their popularity.
We create 3 tasks requiring one entity to have 2 (\textbf{P-2}), 5 (\textbf{P-5}) and 10 (\textbf{P-10}) times the number of links as its partner, and one unrestricted task (\textbf{P-Any}).

\subsection{Factual Knowledge}

Finally, we explore a small set of factual knowledge probes for spatial, temporal, and numeric information using triples of literals from DBPedia.
The first two tasks probe if the embeddings retain the century or decade that a given person was born, based on the embedding for that person (\textbf{F-C} and \textbf{F-D} respectively).
The next two tasks take as input a pair of location-type entities to see if the model can predict which of the two entities is larger in terms of 1) area in square kilometers (\textbf{F-A}) and 2) population (\textbf{F-P}).
We select pairs using two methods, one which compares random pairs of entities and one that uses our type-restricted selection method from above to prevent the model from learning easy, type-based comparisons between, for instance, countries and villages, referring to the type-restricted versions as \textbf{F-A+T} and \textbf{F-P+T}.
The final task compares two organisation type entities and tries to predict which has the higher revenue (\textbf{F-R}).
We restrict pairs in this task to those whose revenues are reported in the same currency.

\subsection{Probing Experiments}

For all tasks, we create train and test sets with 500 entities per label.
For \textbf{R-C+I}, we include 100 corrupted examples from each relationship type, for 24,400 None-type instances.
We use relatively small training sizes and a logistic regression classifier as the probing model to observe how easily the information can be identified from a limited sample and simple model \cite{hewitt2019control}.
For the popularity regression task, we use 500 training and test instances and a linear model trained with Huber loss.
For single entity probing tasks, the input is the embedding of the entity in question.
For tasks probing a pair of entities, the input is the concatenation of the two entities' embeddings, \textit{h} and \textit{t}, as well as $h-t$ and the element-wise product $h \odot t$.
We report macro F1 for all tasks except binary relation detection and context word prediction, where we report macro F1 averaged over all sub-tasks, and the popularity regression task where we report RMSE.

\begin{table*}[ht]
\begin{center}
\tiny
\begin{tabular}{@{}ccc|ccc|cccc|cccccc@{}}
\toprule
Task Category & \multicolumn{2}{c}{\textbf{Words}} & \multicolumn{3}{c}{\textbf{Entity Types}} & \multicolumn{4}{c}{\textbf{Relationships}} & \multicolumn{5}{c}{\textbf{Popularity}} \\ 
\cmidrule(lr){2-3}  \cmidrule(lr){4-6} \cmidrule(lr){7-10} \cmidrule(lr){11-16}
Task Name     & \bf W-H          & \bf W-M         & \bf T-1       & \bf T-2       & \bf T-3      & \bf R-D    & \bf R-I    & \bf R-C    & \bf  R-C+I & \bf P-R & \bf P-B & \bf P-Any & \bf P-2 & \bf P-5 & \bf P-10 \\
\# Labels     & 2            & 2           & 20        & 33        & 60        & 2      & 2      & 244    & 245 & N/A & 4 & 2 & 2 & 2 & 2   \\ \midrule
CNN           & 79.1        & 80.8        & 91.4      & 88.5      & 85.5      & 65.5   & 80.4 &    54.2  & 47.8 & 1.05 & 48.5 & 60.6 & 69.3 & 76.9 & 81.7    \\
RNN           & 57.6           & 56.6        & 6.0       & 3.3       & 2.2       & 55.3     & 63.9   & 16.3   & 16.7  & 1.04 & 55.8 & 60.6  & 73.4   & \textbf{90.2} & \textbf{95.6}  \\
Ganea         & \textbf{86.9}         & \textbf{88.8}        & 93.1      & 89.0        & 84.1      & 60.7     & 75.0     & 72.4   & 67.8 & 1.03 & 60.3 & \textbf{68.1}    & \textbf{80.1} & 87.2   & 95.0  \\
BigGraph      & 70.7         & 73.4        & \textbf{100.0}~       & \textbf{100.0}~       & \textbf{97.5}      & \textbf{81.9}   & \textbf{93.8}   & \textbf{82.0}     & \textbf{81.6} & 1.21 & 43.4  & 56.9  & 59.2 & 68.1 & 78.8  \\
Wiki2V      & 81.5         & 84.5        & 94.8      & 88.7      & 84.8      & 80.0     & 89.8   & 72.1   & 69.1 & \textbf{0.84} & \textbf{62.6} & 66.5  & 75.8 & 84.1   & 87.3  \\
BOW           & 78.9         & 81.2        & 93.6      & 86.8      & 81.1      & 57.2   & 75.9   & 65.4   & 59.2 & 1.05 & 46.9 & 56.2  & 60.9 & 68.1 & 72.9   \\
BERT          & 84.5         & 86.5        & 97.8      & 94.6      & 91.5      & 67.8   & 83.2   & 77.8   & 74.1 & 0.98 & 57.8 & 66.2    & 73.4 & 84.8 & 90.8  \\
BERT-Large    & 84.6      & 86.6        & 97.9      & 94.5      & 91.5      & 66.7   & 82.4   & 78.1   & 74.6 & 0.97 & 57.6 & 65.1  & 75.8 & 85.1 & 92.7       \\ \bottomrule
\end{tabular}
\end{center}
\caption{Results for context word, entity type, relationship and popularity probing tasks. All values are micro F1, except for \textbf{W-H}, \textbf{W-M}, and \textbf{R-I}, which report average macro F1 across all subtasks, and \textbf{P-R} which reports RMSE.}
\label{tab:probe-type}
\end{table*}

\section{Probing Experiment Analysis}
We present the results of our probing tasks in Tables \ref{tab:probe-type} and \ref{tab:probe-fact} and analyze them in the following sections.

\paragraph{Context Word Prediction.} 

\textbf{Ganea} performs best on the two word context tasks, beating even \textbf{BERT} and \textbf{BERT-large}, and demonstrating one of the advantages of their shared word and entity embedding space.
\textbf{CNN} performs on par with the \textbf{BOW} model, indicating that the task-learned embeddings store a fair amount of lexical information to complete the EL task.
The \textbf{RNN} embeddings perform at almost chance level, which could mean the lexical information is stored in the RNN layers and not transferred to the entity embeddings.

Examining the highest performing words across the models, the most common high-performing categories are domain-specific terms like \textit{nhl}, \textit{genus}, and \textit{manga}, place names and demonyms like \textit{china} and \textit{australian}, and entity type descriptors like \textit{rapper} and \textit{pitcher}.
The higher performance on domain-specific words can also be seen in the general increased performance on mid-frequency words, which are often more domain-specific.
This helps explain the surprisingly decent performance of \textbf{BigGraph}, which was not trained with text data, but may be able to use the fine-grained entity type data it was trained on to identify specific domains.

\paragraph{Entity Type Classification.} 

In Table \ref{tab:probe-type}, we see that \textbf{BigGraph} almost perfectly identifies the type system, which we might expect since it is trained in part on fine-grained entity type links.
Even without explicit training on type information, all models except the \textbf{RNN} perform exceptionally well on the entity type prediction task.
Given the relatively small decrease in performance as we increase the granularity of the type set, we expect these results to hold even in larger type sets such as FIGER \cite{ling2020relic}.
Wikipedia pages often start with a sentence like ``Entity X is a Y", where Y contains fine-grained type information about X, leading to strong performance for description-based models like \textbf{CNN} and \textbf{BERT-Large}.
Interestingly, \textbf{Ganea} performs relatively poorly on this task compared to the other models, which could be because it is trained only on context around the entities, and doesn't have direct access to the rich description of the entity.
\textbf{Wiki2V}, which is similarly context-based, is also informed by its links to other entities which may provide additional information as we see in the next sections.
The \textbf{RNN}'s poor performance was also observed by  \citet{aina2019entity}, who saw low accuracy when probing an ``entity-centric" RNN model for entity type information.

\paragraph{Relation Detection.}

For this task, \textbf{BigGraph} and \textbf{Wiki2V} perform best, which is reasonable as they were both trained explicitly with link prediction tasks.
The remaining models perform fairly poorly, though \textbf{CNN}, \textbf{BERT}, and \textbf{BERT-Large} still perform reasonably above chance.
The results of this task will primarily be useful to contextualize the results of the remaining relationship tasks.

\paragraph{Binary Relation Identification.}

On relation identification, we see similar results as on relation detection, though average performance is increased.
While strong performance from \textbf{BigGraph} and \textbf{Wiki2V} may be expected, high scores by models like \textbf{BERT} or \textbf{CNN}, which had no explicit training on relationships and performed poorly on relation detection, prompt further examination.

Some of the best performing tasks ($>$90 F1) for these models feature a less common entity as the head and a more frequent entity as the tail, such as biological classifications (e.g. \textit{kingdom} and \textit{phylum}) and location-related relationships (e.g. \textit{country}, \textit{state}).
Because the Wikipedia knowledge graph is incomplete, certain entities are over-represented making some relationships easy to classify as we see in the \textit{daylightSavingTimeZone} task, where the North American Central Time entity is used in almost half the positive instances.
For other many-to-few relationships like \textit{country}, \textit{state}, and \textit{phylum}, the models may be able to identify corruptions based on the replacement of a high frequency entity with a lower frequency one to get high accuracy.
For 1-1 or 1-few relationships such as \textit{child}, \textit{formerTeam}, and \textit{album}, all models except \textbf{Wiki2V} and \textbf{BigGraph} perform much worse.

\paragraph{Relation Type Classification.}

Relationship classification shows fairly strong results for a difficult task, particularly compared to the general relation detection task.
\textbf{BigGraph}, which was trained to represent each relation type separately, performs best, but \textbf{Ganea}, \textbf{BERT}, and \textbf{BERT-Large} each perform quite well.
\textbf{Wiki2V}, which performed well on the detection and identification tasks, performs worse than these three models, particularly in comparison to the \textbf{BERT} models.

To better understand this, we look at the relation classification + identification task, where all models except \textbf{BigGraph} drop noticeably in performance.
We argue that \textbf{BigGraph} is largely unaffected because it actually encodes the relationships between entities, while the other models rely, at least in part, on type information.
Certain relationship types are easier to classify than others due to the fine-grained types of the entities involved, such as \textit{militaryBranch} or \textit{diocese}.
Identifying these relationships based on entity type would be easy, but introducing negative examples with matching fine-grained entity types will harm performance much more for models that primarily rely on type information.
In the confusion matrices for the \textbf{R-C+I} task, we see a high number of false positives for the None label: \textbf{Ganea} has an average of 56.2 false positives with the None label per relationship type, while \textbf{BERT} has 24.3 and \textbf{BERT-Large} has 23.1.
\textbf{Wiki2V} performs better with 16.1 while \textbf{BigGraph} has only 2.5, demonstrating it can both identify and label relationships.

\begin{figure}
    \centering
    \includegraphics[width=1.0\linewidth]{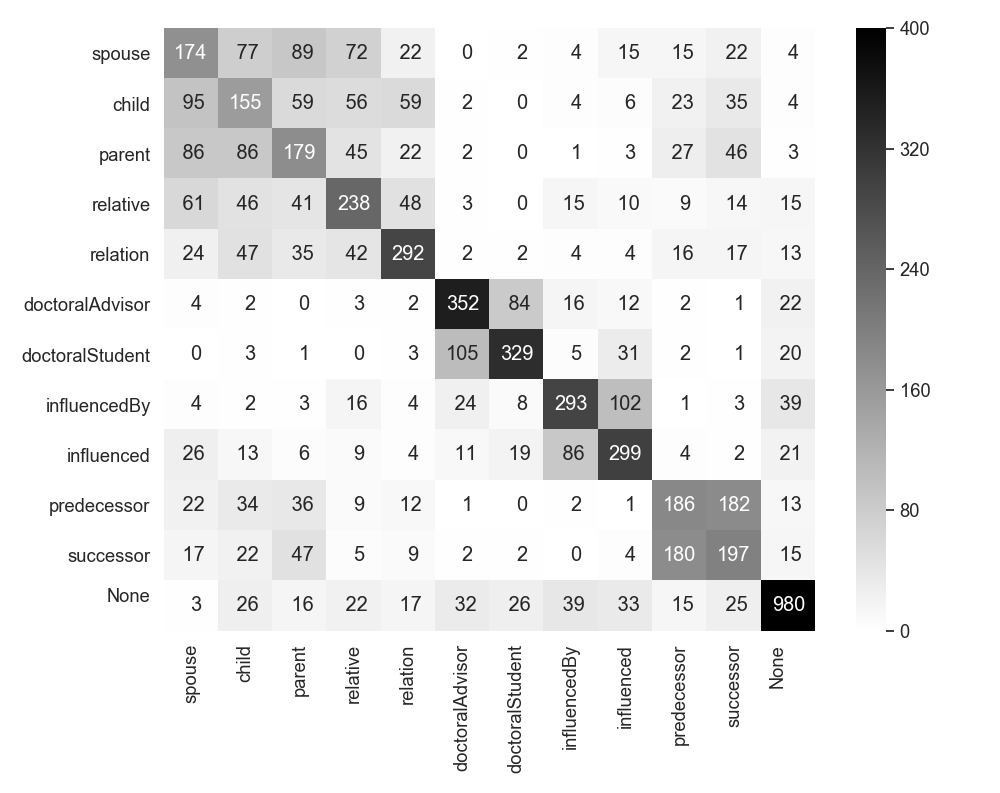}
    \caption{Person-Person relationship confusion matrix for \textbf{Wiki2V}}
    \label{fig:per_conf}
\end{figure}

We next look at small groups of relationships between common entity types to further examine what the high-performing models encode.
Figure \ref{fig:per_conf} shows a confusion matrix from \textbf{Wiki2V} for relationship types between two Person-type entities.

We see one challenge for the model is relationship granularity.
The cluster of family relationships in the top left indicate that the model can generally identify a family relationship, but has difficulty determining the fine-grained label.
\textbf{BigGraph} also makes mistakes on these types, which could indicate label-internal confusion on some types, such as ``relative" and ``relation", whose differences are not apparent or explained in DBPedia.

The second challenge is the relationship direction.
Pairs of relations such as \textit{influenced} and \textit{influencedBy} or \textit{predecessor} and \textit{successor} have high confusion with each other but low confusion with other types.
We see similar trends in relationship pairs such as 
\textit{bandMember} and \textit{formerBandMember} or \textit{parentCompany} and \textit{subsidiary}.
\textbf{Wiki2V} is trained only on binary link prediction, but not the direction.
\textbf{BERT}, which wasn't trained with relationship data, might remember the entity names and related words from pretraining, but not the exact way it was represented, for example if active or passive voice was used.
\textbf{BigGraph} has similar challenges, which could indicate that while the general relationship is expressible with a linear combination of the entities, the direction is not.

\paragraph{Popularity.}

Table \ref{tab:probe-type} shows that \textbf{Wiki2V} performs best on both the popularity regression task and the binned popularity task, particularly outperforming the other models on the regression task.
For the regression task, a simple baseline predicting the average of the training label values results in an RMSE of 1.15.
As such, only \textbf{Wiki2V} actually performs notably better than the baseline, with \textbf{BigGraph} performing even worse than it.
This notable improvement could be due to \textbf{Wiki2V}'s link prediction task, which likely benefits from encoding popularity information as a prior probability that a given entity is linked to by another.

The primary source of errors in the regression task is the highly-linked outliers as might be expected across all the models, but there is no clear consistency across models as far as what types or broad categories of entities seem to be more easily predicted in terms of popularity.
For the binned popularity, the models consistently perform best at identifying entities in the most ($>1000$) and least ($1-10$) popular bins, with most errors coming from entities in the $10-100$ bin and entities whose popularity values are on the edges of a bin.

On the comparative tasks, as the gap in popularity grows, performance increases for all models, supporting our theory that popularity information is mostly coarsely retained.
\textbf{Ganea} overtakes \textbf{Wiki2V} on all comparative tasks, while \textbf{BERT} and \textbf{BERT-Large} only surpass it on the coarser \textbf{P-5} and \textbf{P-10} tasks.
\textbf{Ganea} and the \textbf{BERT} models' high F1 also reinforce our theory that these models can use popularity information to identify true instances of 1-many and many-1 relationships in the \textbf{R-I} task, and it likely also helps \textbf{Wiki2V} given its performance here.
We also finally see a positive result for the \textbf{RNN} model, which has learned coarse differences in entity frequency better than any other model.
Overall we find that many approaches encode coarse, relative popularity information quite well, with \textbf{Ganea} able to detect more fine-grained differences, but they struggle to reproduce accurate estimates of that information.

\begin{table}[ht]
\begin{center}
\tiny
\begin{tabular}{@{}cccccccc@{}}
\toprule
Task Category & \multicolumn{7}{c}{\textbf{Factual}}                      \\ 
\cmidrule(lr){2-8}
Task Name    & \bf F-A & \bf F-A+T & \bf F-P  & \bf F-P+T & \bf F-R & \bf F-C & \bf F-D \\
\# Labels    & 2    & 2     & 2    & 2     & 2    & 5    & 20  \\ \midrule
CNN          & 67.3 & 61.8  & 70.9 & 68.1  & 62.2 & 65.5 & 20.3 \\
RNN          & 54.8 & 48.6  & 56.3 & 53.3  & 51.6 & 21.4 & 5.7  \\
Ganea        & 70.8 & 65.1  & 73.9 & 72.4  & \textbf{71.4} & 62.4 & 20.9 \\
BigGraph     & 66.4 & 55.1  & 65.5 & 62.3  & 58.1 & 44.3 & 13.7 \\
Wiki2V       & 68.1 & 60.1  & 74.0 & 69.3  & 70.7 & 89.4 & 37.4 \\
BOW          & 61.6 & 58.7  & 64.9 & 57.3  & 57.6 & 57.0 & 20.8 \\
BERT         & 75.4 & 67.6  & 79.0 & 74.1  & 68.1 & 90.3  & 45.0 \\
BERT-Large   & \textbf{76.1} & \textbf{70.8}  & \textbf{79.4} & \textbf{76.0}  & 69.8 & \textbf{91.7} & \textbf{48.8} \\ \bottomrule
\end{tabular}
\caption{Results for the factual probing tasks.}
\label{tab:probe-fact}
\end{center}
\end{table}

\paragraph{Factual Knowledge}

In Table \ref{tab:probe-fact} we see that, similar to the relationship classification task, our type-based selection policy for the population and area tasks causes a significant performance drop on otherwise impressive scores, indicating high reliance on type information rather than factual knowledge to predict which entity is larger. 
Performance on the type-restricted tasks also correlates fairly well with performance on the coarse comparative popularity tasks.
Upon closer examination of the four datasets, in 60-70\% of the entity pairs in train and test the first entity (the larger of the two) was also more frequently linked to than the second.
As such, \textbf{BERT}, \textbf{BERT-Large}, and \textbf{Ganea}, which all performed best on the coarser popularity tasks, may have used this information to help on these factual tasks.
The models perform similarly on the revenue task, which has a 65\% rate of the entity with a larger revenue having a higher popularity.

For birth century and decade, \textbf{Wiki2V} and the two \textbf{BERT}-based models perform best.
\textbf{Wiki2V} has a combination of strong context word memory and knowledge of links from other entities, providing a network of contemporary entities that can help narrow down the options.
\textbf{BERT} and \textbf{BERT-Large} have access to the description which often contains a birth year that they can encode directly, yielding higher performance on birth decade prediction than any other model.
While \textbf{BOW} and \textbf{CNN} also have access to this information, their limited vocabularies map most numbers to a generic unknown term so they can only rely on words.

\section{Downstream Task - Entity Linking}

\subsection{Experiments}

\begin{table*}[ht]
\tiny
\begin{tabular}{c|ccc|ccc|ccc}
\toprule
\multirow{2}{*}{Embedding}     & \multicolumn{3}{c|}{\bf CNN}                                                                    & \multicolumn{3}{c|}{\bf RNN}             & \multicolumn{3}{c}{\bf Transformer}                                                                \\
            & \multicolumn{1}{c}{AIDA-Micro} & \multicolumn{1}{c}{AIDA-Macro} & TAC-Micro                 & AIDA-Micro & AIDA-Macro & TAC-Micro & \multicolumn{1}{c}{AIDA-Micro} & \multicolumn{1}{c}{AIDA-Macro} & \multicolumn{1}{c}{TAC-Micro} \\ \midrule
None        & 88.4 & 88.9 & 75.1 & 78.5 & 79.7 & 49.2 & 76.2 & 76.6 & 50.5 \\ 
CNN         & 89.8 & 89.3 & 71.8 & 85.9 & 85.4 & 59.8 & 89.7 & 88.4 & 72.4 \\ 
RNN         & 88.4 & 88.8 & 72.5 & 84.8 & 86.6 & 74.0 & 84.0 & 85.5 & 72.5 \\ 
Ganea       & 90.9 & 90.4 & 77.3 & 87.7 & 88.7 & 78.1 & \textbf{94.2} & \textbf{94.1} & 82.0 \\ 
BigGraph    & 89.4 & 89.5 & 72.3 & 87.0 & 87.4 & 75.6 & 89.6 & 90.1 & 79.5 \\ 
Wiki2V    & 91.2 & 91.2 & 77.9 & \textbf{90.6} & \textbf{90.3} & 78.4 & 92.6 & 92.7 & 80.9 \\ 
BoW         & 89.8 & 89.5 & 71.4 & 87.2 & 88.2 & 72.5 & 92.0 & 92.6 & 74.0 \\ 
BERT-base   & \textbf{91.6} & \textbf{91.4} & 74.8 & 89.0 & 89.8 & 80.2 & 92.0 & 92.7 & 82.8 \\ 
BERT-large  & 90.9 & 91.3 & \textbf{78.9} & 88.4 & 89.3 & \textbf{81.2} & 91.8 & 92.4 & \textbf{83.0} \\
\bottomrule

\end{tabular}
\caption{Entity linking performance of 3 models both without pretrained embeddings and using each of our 8 entity embedding methods. Best values for each model and dataset are in bold.}
\label{tab:el}
\end{table*}

Many of our embedding methods have been evaluated on EL tasks in prior work, either in a separate model or as full EL models themselves.
However, direct comparison of the impact of of the embeddings on EL performance is confounded by differences in the architectures which leverage the embeddings, as well as difficult to reproduce differences in candidate selection, data preprocessing, and other implementation details.
To address this, we evaluate all of our embeddings in a consistent framework, testing them on two standard datasets in three different EL model architectures to directly compare the contribution of the embeddings to performance on the downstream task and how well they perform across different model architectures.

We test the embeddings using three EL models on two standard EL datasets, the AIDA-CoNLL 2003 dataset \cite{hoffart2011kore} and the TAC-KBP 2010 dataset \cite{ji2010overview}.
Two of our EL models are the CNN and RNN EL models used to generate our task-learned embeddings.
Our third is a transformer model based on the RELIC model of \citet{ling2020relic} that encodes a 128-word context window around the entity mention using uncased \texttt{DistilBERT-base} \cite{sanh2019distilbert}\footnote{https://huggingface.co/distilbert-base-uncased}. 
We compare the embedding of the CLS token in the final layer to a separate entity embedding for each candidate entity using a weighted cosine similarity.
To compare the impact of the entity embeddings, we replace the candidate document convolution in the CNN model or the randomly initialized embeddings in the RNN and transformer models with the pretrained embeddings during training.
Details about dataset preprocessing, candidate selection, and model training can be found in Appendix \ref{sec:el_app}.

\subsection{Entity Linking Results}

Table \ref{tab:el} contains the results of our 3 EL models using each of our 8 embedding methods, as well as no pretrained embeddings for comparison.
We report micro-averaged and macro-averaged Precision@1 for AIDA-CoNLL and micro-averaged Precision@1 for TAC-KBP, following previous work \cite{yamada2017texts,eshel2017noisy,raiman2018deeptype}.
Each result is the average of three runs for that configuration.

We see clear benefits from pretrained embeddings across all models and datasets.
While the \textbf{CNN} and \textbf{RNN} embeddings provide improvements compared to using no pretrained embeddings, they transfer poorly to other models, often performing worse than even the simple \textbf{BOW} embedding approach.
\textbf{BigGraph} performs worse than \textbf{BOW} on AIDA-CoNLL, but better on TAC-KBP, which could indicate its strong type information helps more on the smaller dataset, while word information may be more helpful on AIDA.

While the transformer EL model clearly outperforms the CNN and RNN EL models, no single embedding model performs consistently better across these datasets and models.
\textbf{Wiki2V} shows the highest potential for generalization across models on CoNLL-AIDA, possibly because of its combination of context word, entity type, and popularity information, the latter of which has been shown to set a non-trivial baseline on this dataset \cite{chen2019enteval}.
\textbf{Ganea} performs extremely well in combination with the Transformer model, approaching the current state of the art on AIDA-CoNLL set by \citet{raiman2018deeptype}.
\textbf{BERT-Large} consistently performs best on TAC-KBP, a smaller dataset which, as noted above, may benefit more from this model's well-encoded entity type information, and likely also its very strong context word knowledge.
Tasks like knowledge base completion or question answering will require additional information and our probing task results may provide guidance for selecting embeddings for those tasks.

\section{Related Work}

\subsection{Probing Tasks}

Interpretation of neural language representations has drawn increased attention in recent years, particularly with the rise of BERT and transformer-based language models \cite{lipton2018mythos,belinkov2019analysis,tenney2019classic,liu2019transfer}. 
We focus on methods for detecting specific attributes in learned representations, referred to as point-based intrinsic evaluations \cite{yaghoobzadeh2016intrinsic}, auxiliary prediction tasks \cite{adi2017embed} or probing tasks \cite{conneau2018vector,najoung2019probing}.
In these tasks, a model's weights are frozen after training and queried for linguistic knowledge using small classification tasks.

These techniques have similarly been applied to entity embeddings, though usually to limited extents.
Entity type prediction has been among the most common task explored when proposing a new entity embedding method, in part because fine-grained entity type prediction is a common standalone task itself \cite{ling2012fine,gupta2017joint,yaghoobzadeh2017multi,aina2019entity,chen2020type}.
Recently, BERT-inspired techniques have been used to probe entity knowledge stored in pretrained language models through Cloze-style tasks, in which part of a fact about an entity is obscured and the model predicts the missing word(s)  \cite{peters2019know,petroni2019language,poerner2019bert,fevry2020experts}. 
These have yielded tremendous insights, but are limited to models which can directly encode language about an entity and generate new text.
Concurrent with this work, \citet{chen2019enteval} introduced EntEval, a series of probing tasks for both fixed (description-based) and contextual entity embeddings to evaluate semantic type and relationship information in BERT and ELMo-based entity embeddings.
However, like the Cloze-style tasks, many of their tasks are limited to one type of embedding or another and they compare only a small number of unique entity embedding methods while providing limited analysis of the task results.

Our work builds on this prior work, which has often limited either its task exploration, the models being evaluated, or the extent of its analysis.
We propose a set of tasks including several which are, to the best of our knowledge, novel to analyzing entity embeddings such as popularity prediction and context word evaluation.
These tasks can be easily applied to any method which produces a single embedding per entity allowing us to compare a much wider range of model architectures than in any prior work.
Additionally, we provide extensive analysis of performance and errors on these tasks and demonstrate the importance of carefully designing these tasks to better ascertain the true knowledge captured by the embeddings.

\subsection{Neural Entity Linking}
Early neural EL models learned representations by maximizing the similarity between the KB candidate's text and the mention's context \cite{he2013representation,francislandau2016convolutional}.
Approaches based on skip-gram and CBOW models \cite{Mikolov2013word2vec} jointly trained word and entity embeddings, producing state of the art results on EL \cite{yamada2016words,cao2017mpeme,chen2018bilinear}, named entity recognition \cite{sato2017segment}, and question answering \cite{yamada2017texts}.
Some neural EL systems have explicitly included semantic information such as an entity's type \cite{huang2015kg,gupta2017joint,onoe2020interpreTable,chen2020type}.
Recent approaches have explored integrating BERT with pretrained entity embeddings  \cite{zhang2019ernie,peters2019know,poerner2019bert}, while others have used BERT directly to learn entity embeddings for the task \cite{ling2020relic,wang2020kepler,broscheit2020entity}.

\section{Conclusion}

In this work, we propose a new set of probing tasks for evaluating entity embeddings which can be applied to any method that creates one embedding per entity.
Using these tasks, we find that entity type information is one of the strongest signals present in all but one of the embedding models, followed by coarse information about how likely an entity is to be mentioned.
We show that the embeddings are particularly able to use entity type information to bootstrap their way to improved performance on entity relationship and factual information prediction tasks and propose methods to counteract this to more accurately estimate how well they encode relationships and facts.

Overall, we find that while BERT-based entity embeddings perform well on many of these tasks, their high performance can often be attributed to strong entity type information encoding.
More specialized models such as Wikipedia2Vec are better able to detect and identify relationships, while the embeddings of \citet{ganea2017joint} better capture the lexical and distributional semantics of entities.
Additionally, we provide a direct comparison of the embeddings on two downstream EL tasks, where the models that performed well on the probing tasks such as \textbf{Ganea}, \textbf{Wiki2V}, and \textbf{BERT} performed best on the downstream tasks.
We find that the best performing embedding model depends greatly on the surrounding architecture and encourage future practitioners to directly compare newly proposed methods with prior models in a consistent architecture, rather than only compare results.

Our work provides insight into the information encoded by static entity embeddings, but entities can change over time, sometimes quite significantly.
One future line of work we would like to pursue using our tests is to investigate how changes in entities over time can be reflected in the embeddings, and how those changes could be modeled as transformations in the embedding space.
Context-based embeddings in particular could then be dynamically updated with new information, instead of being retrained from scratch.

\bibliographystyle{acl_natbib}
\bibliography{emnlp2020}

\begin{thebibliography}{48}
\expandafter\ifx\csname natexlab\endcsname\relax\def\natexlab#1{#1}\fi

\bibitem[{Adi et~al.(2017)Adi, Kermany, Belinkov, Lavi, and
  Goldberg}]{adi2017embed}
Yossi Adi, Einat Kermany, Yonatan Belinkov, Ofer Lavi, and Yoav Goldberg. 2017.
\newblock \href {https://openreview.net/forum?id=BJh6Ztuxl} {Fine-grained
  analysis of sentence embeddings using auxiliary prediction tasks}.
\newblock In \emph{5th International Conference on Learning Representations,
  {ICLR} 2017, Toulon, France, April 24-26, 2017, Conference Track
  Proceedings}.

\bibitem[{Aina et~al.(2019)Aina, Silberer, Sorodoc, Westera, and
  Boleda}]{aina2019entity}
Laura Aina, Carina Silberer, Ionut{-}Teodor Sorodoc, Matthijs Westera, and
  Gemma Boleda. 2019.
\newblock \href {https://doi.org/10.18653/v1/n19-1378} {What do entity-centric
  models learn? insights from entity linking in multi-party dialogue}.
\newblock In \emph{Proceedings of the 2019 Conference of the North American
  Chapter of the Association for Computational Linguistics: Human Language
  Technologies, {NAACL-HLT} 2019, Minneapolis, MN, USA, June 2-7, 2019, Volume
  1 (Long and Short Papers)}, pages 3772--3783. Association for Computational
  Linguistics.

\bibitem[{Belinkov and Glass(2019)}]{belinkov2019analysis}
Yonatan Belinkov and James~R. Glass. 2019.
\newblock \href {https://transacl.org/ojs/index.php/tacl/article/view/1570}
  {Analysis methods in neural language processing: {A} survey}.
\newblock \emph{Trans. Assoc. Comput. Linguistics}, 7:49--72.

\bibitem[{Broscheit(2019)}]{broscheit2020entity}
Samuel Broscheit. 2019.
\newblock \href {https://doi.org/10.18653/v1/K19-1063} {Investigating entity
  knowledge in {BERT} with simple neural end-to-end entity linking}.
\newblock In \emph{Proceedings of the 23rd Conference on Computational Natural
  Language Learning, CoNLL 2019, Hong Kong, China, November 3-4, 2019}, pages
  677--685. Association for Computational Linguistics.

\bibitem[{Cao et~al.(2017)Cao, Huang, Ji, Chen, and Li}]{cao2017mpeme}
Yixin Cao, Lifu Huang, Heng Ji, Xu~Chen, and Juanzi Li. 2017.
\newblock \href {https://doi.org/10.18653/v1/P17-1149} {Bridge text and
  knowledge by learning multi-prototype entity mention embedding}.
\newblock In \emph{Proceedings of the 55th Annual Meeting of the Association
  for Computational Linguistics, {ACL} 2017, Vancouver, Canada, July 30 -
  August 4, Volume 1: Long Papers}, pages 1623--1633.

\bibitem[{Ceccarelli et~al.(2013)Ceccarelli, Lucchese, Orlando, Perego, and
  Trani}]{ceccarelli2013related}
Diego Ceccarelli, Claudio Lucchese, Salvatore Orlando, Raffaele Perego, and
  Salvatore Trani. 2013.
\newblock \href {https://doi.org/10.1145/2505515.2505711} {Learning relatedness
  measures for entity linking}.
\newblock In \emph{22nd {ACM} International Conference on Information and
  Knowledge Management, CIKM'13, San Francisco, CA, USA, October 27 - November
  1, 2013}, pages 139--148. {ACM}.

\bibitem[{Chen et~al.(2018)Chen, Wei, Liu, Li, Yu, and Zhu}]{chen2018bilinear}
Hui Chen, Baogang Wei, Yonghuai Liu, Yiming Li, Jifang Yu, and Wenhao Zhu.
  2018.
\newblock \href {https://doi.org/10.1016/j.neucom.2017.11.064} {Bilinear joint
  learning of word and entity embeddings for entity linking}.
\newblock \emph{Neurocomputing}, 294:12--18.

\bibitem[{Chen et~al.(2019)Chen, Chu, Chen, Stratos, and
  Gimpel}]{chen2019enteval}
Mingda Chen, Zewei Chu, Yang Chen, Karl Stratos, and Kevin Gimpel. 2019.
\newblock \href {https://doi.org/10.18653/v1/D19-1040} {Enteval: {A} holistic
  evaluation benchmark for entity representations}.
\newblock In \emph{Proceedings of the 2019 Conference on Empirical Methods in
  Natural Language Processing and the 9th International Joint Conference on
  Natural Language Processing, {EMNLP-IJCNLP} 2019, Hong Kong, China, November
  3-7, 2019}, pages 421--433. Association for Computational Linguistics.

\bibitem[{Chen et~al.(2020)Chen, Wang, Jiang, and Lin}]{chen2020type}
Shuang Chen, Jinpeng Wang, Feng Jiang, and Chin{-}Yew Lin. 2020.
\newblock \href {https://aaai.org/ojs/index.php/AAAI/article/view/6251}
  {Improving entity linking by modeling latent entity type information}.
\newblock In \emph{The Thirty-Fourth {AAAI} Conference on Artificial
  Intelligence, {AAAI} 2020, The Thirty-Second Innovative Applications of
  Artificial Intelligence Conference, {IAAI} 2020, The Tenth {AAAI} Symposium
  on Educational Advances in Artificial Intelligence, {EAAI} 2020, New York,
  NY, USA, February 7-12, 2020}, pages 7529--7537. {AAAI} Press.

\bibitem[{Conneau et~al.(2018)Conneau, Kruszewski, Lample, Barrault, and
  Baroni}]{conneau2018vector}
Alexis Conneau, Germ{\'{a}}n Kruszewski, Guillaume Lample, Lo{\"{\i}}c
  Barrault, and Marco Baroni. 2018.
\newblock \href {https://aclanthology.info/papers/P18-1198/p18-1198} {What you
  can cram into a single {\textbackslash}{\textdollar}{\&}!{\#}* vector:
  Probing sentence embeddings for linguistic properties}.
\newblock In \emph{Proceedings of the 56th Annual Meeting of the Association
  for Computational Linguistics, {ACL} 2018, Melbourne, Australia, July 15-20,
  2018, Volume 1: Long Papers}, pages 2126--2136.

\bibitem[{Devlin et~al.(2019)Devlin, Chang, Lee, and
  Toutanova}]{devlin2018bert}
Jacob Devlin, Ming{-}Wei Chang, Kenton Lee, and Kristina Toutanova. 2019.
\newblock \href {https://doi.org/10.18653/v1/n19-1423} {{BERT:} pre-training of
  deep bidirectional transformers for language understanding}.
\newblock In \emph{Proceedings of the 2019 Conference of the North American
  Chapter of the Association for Computational Linguistics: Human Language
  Technologies, {NAACL-HLT} 2019, Minneapolis, MN, USA, June 2-7, 2019, Volume
  1 (Long and Short Papers)}, pages 4171--4186. Association for Computational
  Linguistics.

\bibitem[{Eshel et~al.(2017)Eshel, Cohen, Radinsky, Markovitch, Yamada, and
  Levy}]{eshel2017noisy}
Yotam Eshel, Noam Cohen, Kira Radinsky, Shaul Markovitch, Ikuya Yamada, and
  Omer Levy. 2017.
\newblock \href {https://doi.org/10.18653/v1/K17-1008} {Named entity
  disambiguation for noisy text}.
\newblock In \emph{Proceedings of the 21st Conference on Computational Natural
  Language Learning (CoNLL 2017), Vancouver, Canada, August 3-4, 2017}, pages
  58--68.

\bibitem[{F{\'{e}}vry et~al.(2020)F{\'{e}}vry, Soares, FitzGerald, Choi, and
  Kwiatkowski}]{fevry2020experts}
Thibault F{\'{e}}vry, Livio~Baldini Soares, Nicholas FitzGerald, Eunsol Choi,
  and Tom Kwiatkowski. 2020.
\newblock \href {http://arxiv.org/abs/2004.07202} {Entities as experts: Sparse
  memory access with entity supervision}.
\newblock \emph{CoRR}, abs/2004.07202.

\bibitem[{Francis{-}Landau et~al.(2016)Francis{-}Landau, Durrett, and
  Klein}]{francislandau2016convolutional}
Matthew Francis{-}Landau, Greg Durrett, and Dan Klein. 2016.
\newblock \href {http://aclweb.org/anthology/N/N16/N16-1150.pdf} {Capturing
  semantic similarity for entity linking with convolutional neural networks}.
\newblock In \emph{{NAACL} {HLT} 2016, The 2016 Conference of the North
  American Chapter of the Association for Computational Linguistics: Human
  Language Technologies, San Diego California, USA, June 12-17, 2016}, pages
  1256--1261.

\bibitem[{Ganea and Hofmann(2017)}]{ganea2017joint}
Octavian{-}Eugen Ganea and Thomas Hofmann. 2017.
\newblock \href {https://www.aclweb.org/anthology/D17-1277/} {Deep joint entity
  disambiguation with local neural attention}.
\newblock In \emph{Proceedings of the 2017 Conference on Empirical Methods in
  Natural Language Processing, {EMNLP} 2017, Copenhagen, Denmark, September
  9-11, 2017}, pages 2619--2629.

\bibitem[{Gupta et~al.(2017)Gupta, Singh, and Roth}]{gupta2017joint}
Nitish Gupta, Sameer Singh, and Dan Roth. 2017.
\newblock \href {https://aclanthology.info/papers/D17-1284/d17-1284} {Entity
  linking via joint encoding of types, descriptions, and context}.
\newblock In \emph{Proceedings of the 2017 Conference on Empirical Methods in
  Natural Language Processing, {EMNLP} 2017, Copenhagen, Denmark, September
  9-11, 2017}, pages 2681--2690.

\bibitem[{He et~al.(2013)He, Liu, Li, Zhou, Zhang, and
  Wang}]{he2013representation}
Zhengyan He, Shujie Liu, Mu~Li, Ming Zhou, Longkai Zhang, and Houfeng Wang.
  2013.
\newblock \href {http://aclweb.org/anthology/P/P13/P13-2006.pdf} {Learning
  entity representation for entity disambiguation}.
\newblock In \emph{Proceedings of the 51st Annual Meeting of the Association
  for Computational Linguistics, {ACL} 2013, 4-9 August 2013, Sofia, Bulgaria,
  Volume 2: Short Papers}, pages 30--34.

\bibitem[{Hewitt and Liang(2019)}]{hewitt2019control}
John Hewitt and Percy Liang. 2019.
\newblock \href {https://doi.org/10.18653/v1/D19-1275} {Designing and
  interpreting probes with control tasks}.
\newblock In \emph{Proceedings of the 2019 Conference on Empirical Methods in
  Natural Language Processing and the 9th International Joint Conference on
  Natural Language Processing, {EMNLP-IJCNLP} 2019, Hong Kong, China, November
  3-7, 2019}, pages 2733--2743. Association for Computational Linguistics.

\bibitem[{Hoffart et~al.(2012)Hoffart, Seufert, Nguyen, Theobald, and
  Weikum}]{hoffart2011kore}
Johannes Hoffart, Stephan Seufert, Dat~Ba Nguyen, Martin Theobald, and Gerhard
  Weikum. 2012.
\newblock \href {https://doi.org/10.1145/2396761.2396832} {{KORE:} keyphrase
  overlap relatedness for entity disambiguation}.
\newblock In \emph{21st {ACM} International Conference on Information and
  Knowledge Management, CIKM'12, Maui, HI, USA, October 29 - November 02,
  2012}, pages 545--554.

\bibitem[{Huang et~al.(2015)Huang, Heck, and Ji}]{huang2015kg}
Hongzhao Huang, Larry~P. Heck, and Heng Ji. 2015.
\newblock \href {http://arxiv.org/abs/1504.07678} {Leveraging deep neural
  networks and knowledge graphs for entity disambiguation}.
\newblock \emph{CoRR}, abs/1504.07678.

\bibitem[{Ji et~al.(2010)Ji, Grishman, Dang, Griffitt, and
  Ellis}]{ji2010overview}
Heng Ji, Ralph Grishman, Hoa~Trang Dang, Kira Griffitt, and Joe Ellis. 2010.
\newblock Overview of the tac 2010 knowledge base population track.
\newblock In \emph{Third Text Analysis Conference (TAC 2010)}, volume~3, pages
  3--3.

\bibitem[{Kim et~al.(2019)Kim, Patel, Poliak, Wang, Xia, McCoy, Tenney, Ross,
  Linzen, Durme, Bowman, and Pavlick}]{najoung2019probing}
Najoung Kim, Roma Patel, Adam Poliak, Alex Wang, Patrick Xia, R.~Thomas McCoy,
  Ian Tenney, Alexis Ross, Tal Linzen, Benjamin~Van Durme, Samuel~R. Bowman,
  and Ellie Pavlick. 2019.
\newblock \href {http://arxiv.org/abs/1904.11544} {Probing what different {NLP}
  tasks teach machines about function word comprehension}.
\newblock \emph{CoRR}, abs/1904.11544.

\bibitem[{Kingma and Ba(2015)}]{kingma2015adam}
Diederik~P. Kingma and Jimmy Ba. 2015.
\newblock \href {http://arxiv.org/abs/1412.6980} {Adam: {A} method for
  stochastic optimization}.
\newblock In \emph{3rd International Conference on Learning Representations,
  {ICLR} 2015, San Diego, CA, USA, May 7-9, 2015, Conference Track
  Proceedings}.

\bibitem[{Lerer et~al.(2019)Lerer, Wu, Shen, Lacroix, Wehrstedt, Bose, and
  Peysakhovich}]{lerer2019biggraph}
Adam Lerer, Ledell Wu, Jiajun Shen, Timoth{\'{e}}e Lacroix, Luca Wehrstedt,
  Abhijit Bose, and Alexander Peysakhovich. 2019.
\newblock \href {http://arxiv.org/abs/1903.12287} {Pytorch-biggraph: {A}
  large-scale graph embedding system}.
\newblock \emph{CoRR}, abs/1903.12287.

\bibitem[{Ling et~al.(2020)Ling, FitzGerald, Shan, Soares, F{\'{e}}vry, Weiss,
  and Kwiatkowski}]{ling2020relic}
Jeffrey Ling, Nicholas FitzGerald, Zifei Shan, Livio~Baldini Soares, Thibault
  F{\'{e}}vry, David Weiss, and Tom Kwiatkowski. 2020.
\newblock \href {http://arxiv.org/abs/2001.03765} {Learning cross-context
  entity representations from text}.
\newblock \emph{CoRR}, abs/2001.03765.

\bibitem[{Ling and Weld(2012)}]{ling2012fine}
Xiao Ling and Daniel~S Weld. 2012.
\newblock Fine-grained entity recognition.
\newblock In \emph{AAAI}, volume~12, pages 94--100.

\bibitem[{Lipton(2018)}]{lipton2018mythos}
Zachary~C. Lipton. 2018.
\newblock \href {https://doi.org/10.1145/3233231} {The mythos of model
  interpretability}.
\newblock \emph{Commun. {ACM}}, 61(10):36--43.

\bibitem[{Liu et~al.(2019)Liu, Gardner, Belinkov, Peters, and
  Smith}]{liu2019transfer}
Nelson~F. Liu, Matt Gardner, Yonatan Belinkov, Matthew~E. Peters, and Noah~A.
  Smith. 2019.
\newblock \href {https://doi.org/10.18653/v1/n19-1112} {Linguistic knowledge
  and transferability of contextual representations}.
\newblock In \emph{Proceedings of the 2019 Conference of the North American
  Chapter of the Association for Computational Linguistics: Human Language
  Technologies, {NAACL-HLT} 2019, Minneapolis, MN, USA, June 2-7, 2019, Volume
  1 (Long and Short Papers)}, pages 1073--1094. Association for Computational
  Linguistics.

\bibitem[{Loshchilov and Hutter(2019)}]{loschilov2019adamw}
Ilya Loshchilov and Frank Hutter. 2019.
\newblock \href {https://openreview.net/forum?id=Bkg6RiCqY7} {Decoupled weight
  decay regularization}.
\newblock In \emph{7th International Conference on Learning Representations,
  {ICLR} 2019, New Orleans, LA, USA, May 6-9, 2019}. OpenReview.net.

\bibitem[{Mikolov et~al.(2013)Mikolov, Sutskever, Chen, Corrado, and
  Dean}]{Mikolov2013word2vec}
Tomas Mikolov, Ilya Sutskever, Kai Chen, Gregory~S. Corrado, and Jeffrey Dean.
  2013.
\newblock \href
  {http://papers.nips.cc/paper/5021-distributed-representations-of-words-and-phrases-and-their-compositionality}
  {Distributed representations of words and phrases and their
  compositionality}.
\newblock In \emph{Advances in Neural Information Processing Systems 26: 27th
  Annual Conference on Neural Information Processing Systems 2013. Proceedings
  of a meeting held December 5-8, 2013, Lake Tahoe, Nevada, United States.},
  pages 3111--3119.

\bibitem[{Newman{-}Griffis et~al.(2018)Newman{-}Griffis, Lai, and
  Fosler{-}Lussier}]{newman2018wikisrs}
Denis Newman{-}Griffis, Albert~M. Lai, and Eric Fosler{-}Lussier. 2018.
\newblock \href {https://doi.org/10.18653/v1/w18-3026} {Jointly embedding
  entities and text with distant supervision}.
\newblock In \emph{Proceedings of The Third Workshop on Representation Learning
  for NLP, Rep4NLP@ACL 2018, Melbourne, Australia, July 20, 2018}, pages
  195--206. Association for Computational Linguistics.

\bibitem[{Onoe and Durrett(2020)}]{onoe2020interpreTable}
Yasumasa Onoe and Greg Durrett. 2020.
\newblock \href {http://arxiv.org/abs/2005.00147} {Interpretable entity
  representations through large-scale typing}.
\newblock \emph{CoRR}, abs/2005.00147.

\bibitem[{Pershina et~al.(2015)Pershina, He, and Grishman}]{pershina2015ppr}
Maria Pershina, Yifan He, and Ralph Grishman. 2015.
\newblock \href {http://aclweb.org/anthology/N/N15/N15-1026.pdf} {Personalized
  page rank for named entity disambiguation}.
\newblock In \emph{{NAACL} {HLT} 2015, The 2015 Conference of the North
  American Chapter of the Association for Computational Linguistics: Human
  Language Technologies, Denver, Colorado, USA, May 31 - June 5, 2015}, pages
  238--243.

\bibitem[{Peters et~al.(2019)Peters, Neumann, IV, Schwartz, Joshi, Singh, and
  Smith}]{peters2019know}
Matthew~E. Peters, Mark Neumann, Robert L.~Logan IV, Roy Schwartz, Vidur Joshi,
  Sameer Singh, and Noah~A. Smith. 2019.
\newblock \href {https://doi.org/10.18653/v1/D19-1005} {Knowledge enhanced
  contextual word representations}.
\newblock In \emph{Proceedings of the 2019 Conference on Empirical Methods in
  Natural Language Processing and the 9th International Joint Conference on
  Natural Language Processing, {EMNLP-IJCNLP} 2019, Hong Kong, China, November
  3-7, 2019}, pages 43--54. Association for Computational Linguistics.

\bibitem[{Petroni et~al.(2019)Petroni, Rockt{\"{a}}schel, Riedel, Lewis,
  Bakhtin, Wu, and Miller}]{petroni2019language}
Fabio Petroni, Tim Rockt{\"{a}}schel, Sebastian Riedel, Patrick S.~H. Lewis,
  Anton Bakhtin, Yuxiang Wu, and Alexander~H. Miller. 2019.
\newblock \href {https://doi.org/10.18653/v1/D19-1250} {Language models as
  knowledge bases?}
\newblock In \emph{Proceedings of the 2019 Conference on Empirical Methods in
  Natural Language Processing and the 9th International Joint Conference on
  Natural Language Processing, {EMNLP-IJCNLP} 2019, Hong Kong, China, November
  3-7, 2019}, pages 2463--2473. Association for Computational Linguistics.

\bibitem[{P{\"{o}}rner et~al.(2019)P{\"{o}}rner, Waltinger, and
  Sch{\"{u}}tze}]{poerner2019bert}
Nina P{\"{o}}rner, Ulli Waltinger, and Hinrich Sch{\"{u}}tze. 2019.
\newblock \href {http://arxiv.org/abs/1911.03681} {{BERT} is not a knowledge
  base (yet): Factual knowledge vs. name-based reasoning in unsupervised {QA}}.
\newblock \emph{CoRR}, abs/1911.03681.

\bibitem[{Raiman and Raiman(2018)}]{raiman2018deeptype}
Jonathan Raiman and Olivier Raiman. 2018.
\newblock \href
  {https://www.aaai.org/ocs/index.php/AAAI/AAAI18/paper/view/17148} {Deeptype:
  Multilingual entity linking by neural type system evolution}.
\newblock In \emph{Proceedings of the Thirty-Second {AAAI} Conference on
  Artificial Intelligence, (AAAI-18), the 30th innovative Applications of
  Artificial Intelligence (IAAI-18), and the 8th {AAAI} Symposium on
  Educational Advances in Artificial Intelligence (EAAI-18), New Orleans,
  Louisiana, USA, February 2-7, 2018}, pages 5406--5413.

\bibitem[{Sanh et~al.(2019)Sanh, Debut, Chaumond, and
  Wolf}]{sanh2019distilbert}
Victor Sanh, Lysandre Debut, Julien Chaumond, and Thomas Wolf. 2019.
\newblock \href {http://arxiv.org/abs/1910.01108} {Distilbert, a distilled
  version of {BERT:} smaller, faster, cheaper and lighter}.
\newblock \emph{CoRR}, abs/1910.01108.

\bibitem[{Sato et~al.(2017)Sato, Shindo, Yamada, and
  Matsumoto}]{sato2017segment}
Motoki Sato, Hiroyuki Shindo, Ikuya Yamada, and Yuji Matsumoto. 2017.
\newblock \href {https://aclanthology.info/papers/I17-2017/i17-2017}
  {Segment-level neural conditional random fields for named entity
  recognition}.
\newblock In \emph{Proceedings of the Eighth International Joint Conference on
  Natural Language Processing, {IJCNLP} 2017, Taipei, Taiwan, November 27 -
  December 1, 2017, Volume 2: Short Papers}, pages 97--102.

\bibitem[{Tenney et~al.(2019)Tenney, Das, and Pavlick}]{tenney2019classic}
Ian Tenney, Dipanjan Das, and Ellie Pavlick. 2019.
\newblock \href {https://doi.org/10.18653/v1/p19-1452} {{BERT} rediscovers the
  classical {NLP} pipeline}.
\newblock In \emph{Proceedings of the 57th Conference of the Association for
  Computational Linguistics, {ACL} 2019, Florence, Italy, July 28- August 2,
  2019, Volume 1: Long Papers}, pages 4593--4601. Association for Computational
  Linguistics.

\bibitem[{Wang et~al.(2019)Wang, Gao, Zhu, Liu, Li, and Tang}]{wang2020kepler}
Xiaozhi Wang, Tianyu Gao, Zhaocheng Zhu, Zhiyuan Liu, Juanzi Li, and Jian Tang.
  2019.
\newblock \href {http://arxiv.org/abs/1911.06136} {{KEPLER:} {A} unified model
  for knowledge embedding and pre-trained language representation}.
\newblock \emph{CoRR}, abs/1911.06136.

\bibitem[{Wang et~al.(2014)Wang, Zhang, Feng, and Chen}]{wang2014graph}
Zhen Wang, Jianwen Zhang, Jianlin Feng, and Zheng Chen. 2014.
\newblock \href {http://www.aaai.org/ocs/index.php/AAAI/AAAI14/paper/view/8531}
  {Knowledge graph embedding by translating on hyperplanes}.
\newblock In \emph{Proceedings of the Twenty-Eighth {AAAI} Conference on
  Artificial Intelligence, July 27 -31, 2014, Qu{\'{e}}bec City, Qu{\'{e}}bec,
  Canada.}, pages 1112--1119.

\bibitem[{Yaghoobzadeh and Sch{\"{u}}tze(2016)}]{yaghoobzadeh2016intrinsic}
Yadollah Yaghoobzadeh and Hinrich Sch{\"{u}}tze. 2016.
\newblock Intrinsic subspace evaluation of word embedding representations.
\newblock In \emph{Proceedings of the 54th Annual Meeting of the Association
  for Computational Linguistics, {ACL} 2016, August 7-12, 2016, Berlin,
  Germany, Volume 1: Long Papers}. The Association for Computer Linguistics.

\bibitem[{Yaghoobzadeh and Sch{\"{u}}tze(2017)}]{yaghoobzadeh2017multi}
Yadollah Yaghoobzadeh and Hinrich Sch{\"{u}}tze. 2017.
\newblock \href {https://doi.org/10.18653/v1/e17-1055} {Multi-level
  representations for fine-grained typing of knowledge base entities}.
\newblock In \emph{Proceedings of the 15th Conference of the European Chapter
  of the Association for Computational Linguistics, {EACL} 2017, Valencia,
  Spain, April 3-7, 2017, Volume 1: Long Papers}, pages 578--589. Association
  for Computational Linguistics.

\bibitem[{Yamada et~al.(2018)Yamada, Asai, Shindo, Takeda, and
  Takefuji}]{yamada2018wiki2vec}
Ikuya Yamada, Akari Asai, Hiroyuki Shindo, Hideaki Takeda, and Yoshiyasu
  Takefuji. 2018.
\newblock \href {http://arxiv.org/abs/1812.06280} {Wikipedia2vec: An optimized
  tool for learning embeddings of words and entities from wikipedia}.
\newblock \emph{CoRR}, abs/1812.06280.

\bibitem[{Yamada et~al.(2016)Yamada, Shindo, Takeda, and
  Takefuji}]{yamada2016words}
Ikuya Yamada, Hiroyuki Shindo, Hideaki Takeda, and Yoshiyasu Takefuji. 2016.
\newblock \href {http://aclweb.org/anthology/K/K16/K16-1025.pdf} {Joint
  learning of the embedding of words and entities for named entity
  disambiguation}.
\newblock In \emph{Proceedings of the 20th {SIGNLL} Conference on Computational
  Natural Language Learning, CoNLL 2016, Berlin, Germany, August 11-12, 2016},
  pages 250--259.

\bibitem[{Yamada et~al.(2017)Yamada, Shindo, Takeda, and
  Takefuji}]{yamada2017texts}
Ikuya Yamada, Hiroyuki Shindo, Hideaki Takeda, and Yoshiyasu Takefuji. 2017.
\newblock \href {https://transacl.org/ojs/index.php/tacl/article/view/1065}
  {Learning distributed representations of texts and entities from knowledge
  base}.
\newblock \emph{{TACL}}, 5:397--411.

\bibitem[{Zhang et~al.(2019)Zhang, Han, Liu, Jiang, Sun, and
  Liu}]{zhang2019ernie}
Zhengyan Zhang, Xu~Han, Zhiyuan Liu, Xin Jiang, Maosong Sun, and Qun Liu. 2019.
\newblock \href {https://doi.org/10.18653/v1/p19-1139} {{ERNIE:} enhanced
  language representation with informative entities}.
\newblock In \emph{Proceedings of the 57th Conference of the Association for
  Computational Linguistics, {ACL} 2019, Florence, Italy, July 28- August 2,
  2019, Volume 1: Long Papers}, pages 1441--1451. Association for Computational
  Linguistics.

\end{thebibliography}

\appendix

\section{Entity Linking Task and Model Configuration}
\label{sec:el_app}

\subsection{Data Preprocessing}

As described above, we use two standard entity linking datasets for evaluation, the AIDA-CoNLL 2003 dataset \cite{hoffart2011kore} and the TAC-KBP 2010 dataset \cite{ji2010overview}. 
Following prior work \cite{hoffart2011kore, yamada2017texts}, we evaluate only the mentions that have valid entries in the KB.
TAC-KBP does not have a dedicated validation set, so we assign a random 10\% of the training data to the validation set. 
For candidate set generation for AIDA-CoNLL, we use the PPRforNED candidate sets \cite{pershina2015ppr}.
For TAC-KBP, we pick candidates for each mention that either match the mention, a word in the mention, or have an anchor text that matches the mention.
We keep only the top thirty candidates based on the popularity of the candidate entity in Wikipedia defined as $|M_e|/|M_*|$, where $M_e$ is the number links pointing to the entity and $M_*$ is the total number of links in Wikipedia \cite{yamada2016words}, the same as our definition of popularity in the probing tasks.
For the CNN and RNN models, we lowercase both the KB and mention text, omit stop words and punctuation, and replace all numbers with a single token.
For the transformer model, we only lowercase the text.

\subsection{Model Parameters and Training}

For the CNN model, we use a kernel width of 150 for all convolutions.
To insert pretrained entity embeddings, we replace the candidate document convolution layer, followed by a single layer MLP to reduce the embedding to 150 dimensions.
We apply dropout to the word embedding layer with a probability of 0.2.

For the RNN model, we use single-layer GRUs with hidden size 300 to embed the left and right context around a mention.
We do not tie the weights of the two GRUs.
The MLP attention module takes the candidate entity's embedding as the attention context, applying a linear transform to the embedding to map it to 300 dimensions.
We concatenate the entity embedding to the outputs of the attention module for the left and right contexts and pass the concatenated output to a classifier module consisting of a 300x300 MLP with ReLU activation, followed by a 300x1 linear layer to compute the score.
We apply dropout of 0.2 to the word embeddings and dropout of 0.5 to the MLP in the classification module.
To use pretrained entity embeddings, we replace the randomly initialized entity embeddings with our pretrained embeddings.
In cases where a pretrained entity embedding is unavailable, we randomly initialize the entity's embedding from a uniform distribution between -0.1 and 0.1.
For both the RNN and CNN, we initialize the word embeddings using GoogleNews Word2Vec vectors\footnote{https://code.google.com/archive/p/word2vec}.

For the Transformer model, we use the \texttt{distilbert-base-uncased} model available through the HuggingFace library\footnote{https://huggingface.co/distilbert-base-uncased}.
When using pretrained entity embeddings, we replace the randomly initialized entity embeddings with our pretrained embeddings and randomly initialize any missing entity embeddings from a normal distribution with mean 0 and standard deviation 0.02.
When the pretrained embeddings do not match the 768 dimension output of the DistilBERT context encoder, we map the context encoder's output to match the size of the embeddings with a single linear layer.

All of our models are trained to convergence using hinge loss, with early stopping based on the model's loss on the validation set.
We set our patience for early stopping to 3 epochs for AIDA-CoNLL and 5 for TAC-KBP.
We use batch size 16 for the CNN and RNN models and 32 for the transformer model.
During training, we apply gradient clipping of 1.0 for the transformer model and 5.0 for the CNN and RNN.
We use Adam \cite{kingma2015adam} with an initial learning rate of 1e-3 for the CNN and RNN models, while for the transformer model we use the weight decay-fixed implementation of Adam from HuggingFace \cite{loschilov2019adamw} with initial learning rate of 2e-5 and epsilon 1e-8.
We additionally use a learning rate schedule for the transformer model, with linear decay over the course of training based on an expected maximum number of steps equal to 10 training epochs times the number of batches for the dataset.
When training the CNN and RNN models on the Wikipedia EL corpus, we use all the same model and training settings as described above, but use batch size 512.

\end{document}